\documentclass[11pt,a4paper]{article}
\usepackage[hyperref]{acl2019}
\usepackage{times}
\usepackage{latexsym}

\usepackage{url}

\usepackage{amsmath,amssymb}
\usepackage{algorithm, algorithmicx, algpseudocode} 
\usepackage{graphicx} 
\usepackage{multirow,array}
\usepackage{paralist}
\usepackage{enumitem}
\usepackage{booktabs}

\aclfinalcopy 
\def\aclpaperid{549} 

\title{Budgeted Policy Learning for Task-Oriented Dialogue Systems}

\author{Zhirui Zhang$^\dag$\quad Xiujun Li$^{\ddag\S}$\quad Jianfeng Gao$^\ddag$\quad Enhong Chen$^\dag$\\
  $^\dag$University of Science and Technology of China\\
  $^\ddag$Microsoft Research AI\quad\quad $^\S$University of Washington\\
  \texttt{$^\dag$zrustc11@gmail.com \ \ $^\dag$cheneh@ustc.edu.cn} \\
  \texttt{$^\ddag$\{xiul,jfgao\}@microsoft.com}
}

\date{}

\begin{document}
\maketitle
\begin{abstract}

This paper presents a new approach that extends Deep Dyna-Q (DDQ) by incorporating a Budget-Conscious Scheduling (BCS) to best utilize a fixed, small amount of user interactions (budget) for learning task-oriented dialogue agents. BCS consists of (1) a Poisson-based global scheduler to allocate budget over different stages of training; (2) a controller to decide at each training step whether the agent is trained using real or simulated experiences; (3) a user goal sampling module to generate the experiences that are most effective for policy learning. Experiments on a movie-ticket booking task with simulated and real users show that our approach leads to significant improvements in success rate over the state-of-the-art baselines given the fixed budget.

\end{abstract}

\section{Introduction}
There has been a growing interest in exploiting reinforcement learning (RL) for dialogue policy learning in task-oriented dialogue systems~\cite{levin1997learning,williams2008best,Young2013POMDPBasedSS,Fatemi2016PolicyNW,Zhao2016TowardsEL,Su2016ContinuouslyLN,Li2017EndtoEndTN,Williams2017HybridCN,Dhingra2017EndtoEndRL,Budzianowski2017SubdomainMF,Chang2017AffordableOD,Liu2017IterativePL,Liu2018DialogueLW,gaosurvey}. This is a challenging machine learning task because an RL learner requires real users to interact with a dialogue agent constantly to provide feedback. The process incurs significant real-world cost for complex tasks, such as movie-ticket booking and travel planning, which require exploration in a large state-action space.

In reality, we often need to
develop a dialogue agent with some fixed, limited budget due to limited project funding, conversational data, and development time. Specifically, in this study we measure \emph{budget} in terms of the number of real user interactions. That is, we strive to optimize a dialogue agent via a fixed, small number of interactions with real users.

One common strategy is to leverage a user simulator built on human conversational data \cite{Schatzmann2007AgendaBasedUS,li2016user}. 
However, due to design bias and the limited amounts of publicly available human conversational data for training the simulator, there always exists discrepancies between the behaviors of real and simulated users, which inevitably leads to a sub-optimal dialogue policy. Another strategy is to integrate planning into dialogue policy learning, as the Deep Dyna-Q (DDQ) framework~\cite{Peng2018IntegratingPF}, which effectively leverages a small number of real experiences to learn a dialogue policy efficiently. In DDQ, the limited amounts of real user experiences are utilized for: (1) training a \textit{world model} to mimic real user behaviors and generate simulated experiences; and (2) improving the dialogue policy using both real experiences via direct RL and simulated experiences via indirect RL (planning). 
Recently, some DDQ variants further incorporate discriminators~\cite{Su2018DiscriminativeDD} and active learning~\cite{SwtichBasdDDQYuexin2019} into planning to obtain high-quality simulated experiences.




\begin{figure*}[t]
\centering
\includegraphics[scale=0.5]{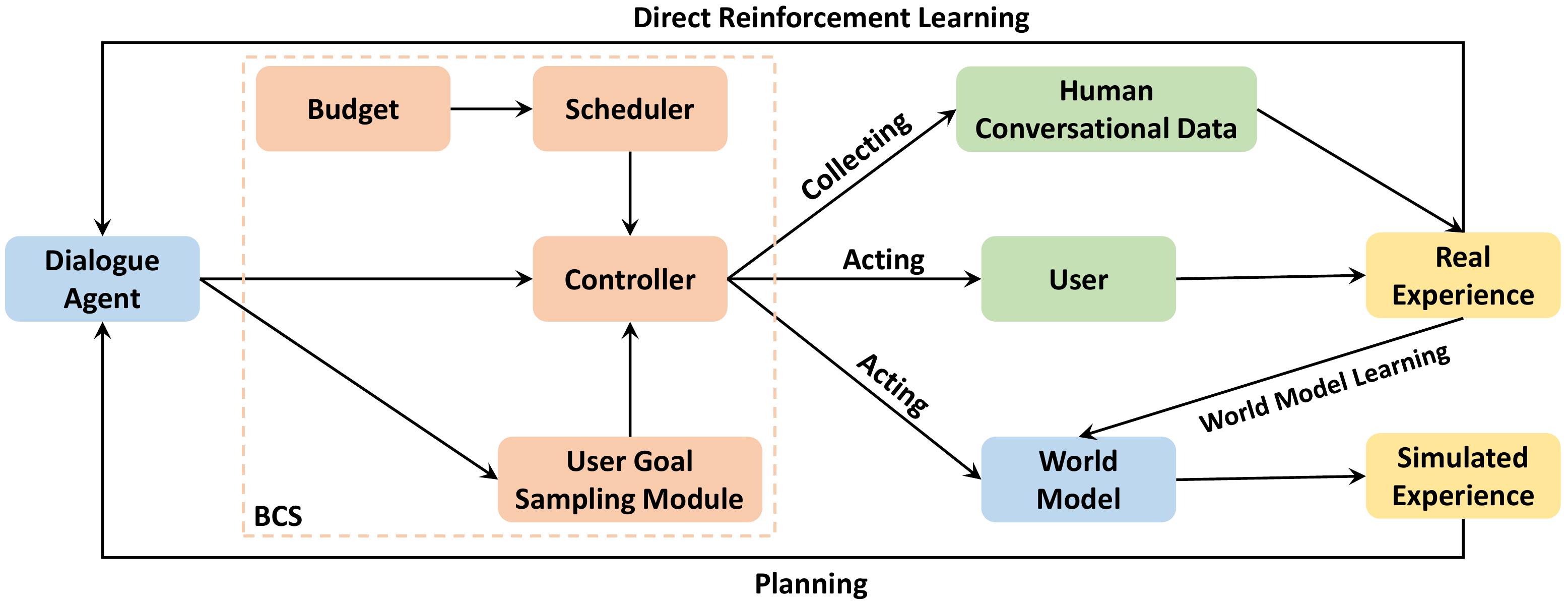}
\caption{Proposed BCS-DDQ framework for dialogue policy learning. BCS represents the Budget-Conscious Scheduling module, which consists of a scheduler, a controller and a user goal sampling module.}
\label{fig:BCS-DDQ-framework}
\end{figure*}



DDQ and its variants face two challenges in the fixed-budget setting. First,  DDQ lacks any explicit guidance on how to generate highly effective real dialogue experiences. For example, the experiences in the state-action space that has not, or less, been explored by the dialogue agent are usually more desirable.
Second, DDQ lacks a mechanism of letting a human (teacher) play the role of the agent to explicitly demonstrate how to drive the dialogue \cite{Barlier2018TrainingDS}. This is useful in the cases where the dialogue agent fails to respond to users in conversations and the sparse negative rewards fail to help the agent improve its dialogue policy. To this end, DDQ needs to be equipped with the ability to decide whether to learn from human demonstrations or from agent-user interactions where the user can be a real user or simulated by the world model.  



In this paper, we propose a new framework, called Budget-Conscious Scheduling-based Deep Dyna-Q (BCS-DDQ), to best utilize a fixed, small number of human interactions (budget) for task-oriented dialogue policy learning. Our new framework extends DDQ by incorporating Budget-Conscious Scheduling (BCS), which aims to control the budget and improve DDQ's sample efficiency by leveraging active learning and human teaching to handle the aforementioned issues. As shown in Figure \ref{fig:BCS-DDQ-framework}, the BCS module consists of (1) a Poisson-based global \textit{scheduler} to allocate budget over the different stages of training; (2) a \textit{user goal sampling module} to select previously failed or unexplored user goals to generate experiences that are effective for dialogue policy learning; (3) a \textit{controller} which decides (based on the pre-allocated budget and the agent's performance on the sampled user goals) whether to collect human-human conversation, or to conduct human-agent interactions to obtain high-quality real experiences, or to generate simulated experiences through interaction with the world model. During dialogue policy learning, real experiences are used to train the world model via supervised learning (world model learning) and directly improve the dialogue policy via direct RL, while simulated experiences are used to enhance the dialogue policy via indirect RL (planning).

Experiments on the movie-ticket booking task with simulated and real users show that our approach leads to significant improvements in success rate over the state-of-the-art baselines given a fixed budget. Our main contributions are two-fold:
\begin{itemize}[noitemsep,leftmargin=*,topsep=0pt]
\item We propose a BCS-DDQ framework, to best utilize a fixed, small amount of user interactions (budget) for task-oriented dialogue policy learning. 
\item We empirically validate the effectiveness of BCS-DDQ on a movie-ticket booking domain with simulated and real users. 
\end{itemize}


\section{Budget-Conscious Scheduling-based Deep Dyna-Q (BCS-DDQ)}

\begin{figure}[t]
\centering
\includegraphics[width=\linewidth]{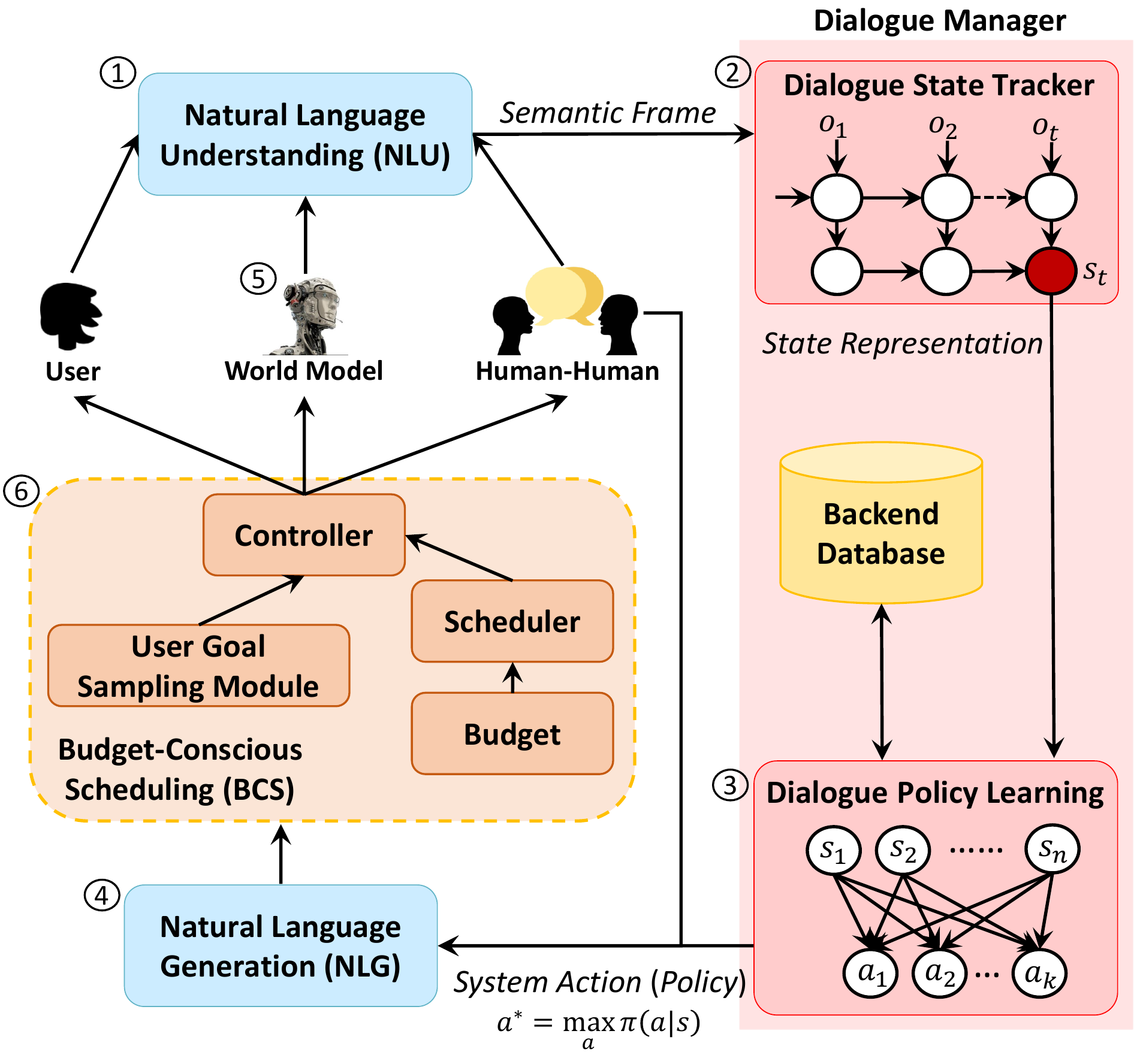}
\caption{Illustration of the proposed BCS-DDQ dialogue system.}
\label{fig:BCS-DDQ-dialogue-system}
\end{figure}

As illustrated in Figure \ref{fig:BCS-DDQ-dialogue-system}, the BCS-DDQ dialogue system consists of six modules: (1) an LSTM-based natural language understanding (NLU) module \cite{HakkaniTr2016MultiDomainJS} for identifying user intents and extracting associated slots; (2) a state tracker \cite{Mrksic2017NeuralBT} for tracking dialogue state; (3) a dialogue policy that chooses the next action based on the current state and database results; (4) a model-based natural language generation (NLG) module for producing a natural language response \cite{Wen2015SemanticallyCL}; (5) a world model for generating simulated user actions and simulated rewards; and (6) the BCS module incorporating a global scheduler, a user goal sampling module and a controller, to manage the budget and select the most effective way to generate real or simulated experiences for learning a dialogue policy.



To leverage BCS in dialogue policy learning, we design a new iterative training algorithm, called BCS-DDQ, as summarized in Algorithm \ref{alg:BCS-DDQ-training}. It starts with an initial dialogue policy and an initial world model, both trained with pre-collected human conversational data. Given the total budget $b$ and maximum number of training epochs $m$, the scheduler allocates the available budget $b_k$ at each training step. Then, the user goal sampling module actively selects a previously failed or unexplored user goal $g_u$. Based on the agent's performance and the current pre-allocated budget, the controller chooses the most effective way, with cost $c^{u} = \{0, 1, 2\}$, to generate real or simulated experiences $B^{r}/B^{s}$ for this sampled user goal. For convenience, the cost of different dialogue generation methods is defined as the number of people involved: 
\begin{itemize}[noitemsep,leftmargin=*,topsep=0pt]
\item cost $c^{u}$ = 2 for collecting human-human demonstrated conversational data.
\item cost $c^{u}$ = 1 for conducting the interactions between human and agent.
\item cost $c^{u}$ = 0 for performing the interactions between world model and agent.
\end{itemize}
The generated experiences are used to update the dialogue policy and the world model.
This process continues until all pre-allocated budget is exhausted. 
In the rest of this section, we detail the components of BCS, and describe the learning methods of the dialogue agent and the world model.
\begin{algorithm}[!htb]
\caption{BCS-DDQ for Dialogue Policy Learning}
\label{alg:BCS-DDQ-training}
\textbf{Input:} The total budget $b$, the maximum number of training epochs $m$, the dialogue agent $A$ and the world model $W$ (both pre-trained with pre-collected human conversational data); 
\begin{algorithmic}[1]
\Procedure{training process}{}
\While{$k < m$}
\State $b_k \leftarrow$ Scheduler($b$, $m$, $k$);
\Repeat
\State $g^{u} \leftarrow$ UserGoalSampler($A$);
\State $B^{r}, B^{s}, c^{u} \leftarrow$ Controller($g^{u}$, $b_k$, $A$, $W$);
\State $b_k \leftarrow b_k - c^{u}$;
\Until {$b_k \leq 0 $}
\State Train the dialogue agent $A$ on $B^{r} \cup B^{s}$
\State Train world model $W$ on $B^{r}$
\EndWhile
\EndProcedure
\end{algorithmic}
\end{algorithm}

\subsection{Budget-Conscious Scheduling (BCS)}
As illustrated in Figure \ref{fig:BCS-DDQ-dialogue-system} and Algorithm \ref{alg:BCS-DDQ-training}, 
BSC consists of a budget allocation algorithm for the scheduler, an active sampling strategy for the user goal sampling module, and a selection policy for the controller.

\subsubsection{Poisson-based Budget Allocation}
The global scheduler is designed to allocate budget ${\{b_1,\dots, b_m \}}$ (where $m$ is the final training epoch) during training. 
The budget allocation process can be viewed as a series of random events, where the allocated budget is a random variable.
In this manner, the whole allocation process essentially is a discrete stochastic process, which can be modeled as a Poisson process. 
Specifically, at each training step $k$, the probability distribution of a random variable $b_k$ equaling $n$ is given by:
\begin{equation}
	P\{b_k=n\} = \frac{\lambda_k^{n}}{n!}e^{-\lambda_k}, \lambda_k = \frac{m+1-k}{m}\lambda
\label{equ:prob-budget-allocation}
\end{equation}
The global scheduling in BCS is based on a \emph{Decayed Possion Process}, motivated by two considerations: 
(1) 
For simplicity, we assume that all budget allocations are mutually-independent. The Poisson process is suitable for this assumption. 
(2) As the training process progresses, the dialogue agent tends to produce higher-quality dialogue experiences using the world model due to the improving performance of both the agent and the world model. As a result, the budget demand for the agent decays during the course of training. 
Thus, we linearly decay the parameter of the Poisson distribution so as to allocate more budget at the beginning of training. 

In addition, to ensure that the sum of the allocated budget does not exceed the total budget $b$, we impose the following constraint:
\begin{equation}
	\sum_{k=1}^m \mathbb{E}[b_k] = \sum_{k=1}^m \frac{m+1-k}{m} \lambda \leq b
\end{equation}
Using this formula, we can calculate the range of the parameter value: $\lambda \leq \frac{2b}{m+1}$. In our experiments, we set $\lambda = \frac{2b}{m+1}$ and sample $b_k$ from the probability distribution defined in Equation \ref{equ:prob-budget-allocation}. 

\subsubsection{Active Sampling Strategy}
The active sampling strategy involves the definition of a user goal space and sampling algorithm.

In a typical task-oriented dialogue \cite{Schatzmann2007AgendaBasedUS}, the user begins a conversation with a user goal $g^{u}$ which consists of multiple constraints. In fact, these constraints correspond to attributes in the knowledge base. For example, in the movie-ticket-booking scenario, the constraints may be the name of the theater (\texttt{theater}), the number of tickets to buy (\texttt{numberofpeople}) or the name of the movie (\texttt{moviename}), and so on. 
Given the knowledge base, we can generate large amounts of user goals by traversing the combination of all the attributes, and then filtering unreasonable user goals which are not similar to real user goals collected from human-human conversational data. We then group the user goals with the same inform and request slots into a category. 
Suppose there are altogether $l$ different categories of user goals. We design a Thompson-Sampling-like algorithm \cite{Chapelle2011AnEE,Eckles2014ThompsonSW,russo2018tutorial} to actively select a previously failed or unexplored user goal in two steps. 
\begin{compactitem}
	\item Draw a number $p_i$ for each category following $ p_i \sim \mathcal{N}(f_i, \sqrt{\frac{l \ln N}{n_i}}) $, where $\mathcal{N}$ represents the Gaussian distribution, $f_i$ denotes the failure rate of each category estimated on the validation set, $n_i$ is the number of samples for each category and $N=\sum_{i} n_i$.
	\item Select the category with maximum $p_i$, then randomly sample a user goal $g_u$ in the category.
\end{compactitem}
Using this method, user goals in the categories with higher failure rates or less exploration are more likely to be selected during training, which encourages the real or simulated user to generate dialogue experiences in the state-action space that the agent has not fully explored.

\subsubsection{Controller}
Given a sampled user goal $g^{u}$, based on the agent's performance on $g^{u}$ and pre-allocated budget $b_k$, the controller decides whether to collect human-human dialogues, human-agent dialogues, or simulated dialogues between the agent and the world model. We design a heuristic selection policy of Equation \ref{equ:controller-budget-allocation} where dialogue experiences $B$ are collected as follow: we first generate a set of simulated dialogues $B^s$ given $g^{u}$, 
and record the success rate $S_{g^{u}}$. If $S_{g^{u}}$ is higher than a threshold $\lambda_1$ (i.e. $\lambda_1 = 2/3$) or there is no budget left, we use $B^s$ for training. If $S_{g^{u}}$ is lower than a threshold $\lambda_2$ (i.e. $\lambda_2=1/3$) and there is still budget, we resort to human agents and real users to generate real experiences $B^{r}_{hh}$. Otherwise, we collect real experiences generated by human users and the dialogue agent $B^{r}_{ha}$.
\begin{align}
(B, c^{u}) = \left\{
\begin{array}{rcl}
 (B^{s}\ \ ,\ 0) & &\text{if}~S_{g^{u}} \geq \lambda_1 ~\text{or}~ b_k = 0 \\
 (B^{r}_{hh},\ 2) & &\text{if}~S_{g^{u}} \leq \lambda_2 ~\text{and}~ b_k \geq 2\\
 (B^{r}_{ha},\ 1) & &\text{otherwise}
\end{array}
\right.
\label{equ:controller-budget-allocation}
\end{align}

Combined with the active sampling strategy, this selection policy makes fuller use of the budget to generate experiences that are most effective for dialogue policy learning.

\subsection{Direct Reinforcement Learning and Planning}
Policy learning in task-oriented dialogue using RL can be cast as a Markov Decision Process 
which consists of a sequence of $<$state, action, reward$>$ tuples. 
We can use the same Q-learning algorithm to train the dialogue agent using either real or simulated experiences.
Here we employ the Deep Q-network (DQN) \cite{Mnih2015HumanlevelCT}.

Specifically, at each step, the agent observes the dialogue state $s$, then chooses an action $a$ using an $\epsilon$-greedy policy that selects a random action with probability $\epsilon$, and otherwise follows the greedy policy $a = \arg\max_{a'} Q(s,a';\theta_{Q})$. $Q(s,a;\theta_{Q})$ approximates the state-action value function with a Multi-Layer Perceptron (MLP) parameterized by $\theta_{Q}$. 
Afterwards, the agent receives reward $r$, observes the next user or simulator response, and updates the state to $s'$. The experience $(s, a, r, a^{u}, s')$ is then stored in a real experience buffer $B^{r}$\footnote{$B^{r}$ = \{$B^{r}_{hh}, B^{r}_{ha}$\}} or simulated experience buffer $B^{s}$ depending on the source. Given these experiences, we optimize the value function $Q(s,a;\theta_{Q})$ through mean-squared loss:
\begin{equation}
\begin{aligned}
\mathcal{L}(\theta_Q) & = \mathbb{E}_{{(s,a,r,s')} \sim B^{r} \cup B^{s}} [(y - Q(s,a;\theta_Q))^2] \\
y & = r + \gamma \max_{a'}Q'(s',a';\theta_{Q'})
\end{aligned}
\end{equation}
where $\gamma \in [0,1]$ is a discount factor, and $Q'(\cdot)$ is the target value function that is updated only periodically (i.e., fixed-target).
The updating of $Q(\cdot)$ thus is conducted through differentiating this objective function via mini-batch gradient descent.

\subsection{World Model Learning}

We utilize the same design of the world model in \citet{Peng2018IntegratingPF}, which is implemented as a multi-task deep neural network. At each turn of a dialogue, the world model takes the current dialogue state $s$ and the last system action $a$ from the agent as input, and generates the corresponding  user response $a^{u}$, reward $r$, and a binary termination signal $t$. 
The computation for each term can be shown as below:
\begin{equation}
\begin{aligned}
h\ & =\ \text{tanh}(W_h [s,a] + b_h) \\
r\ & =\ W_r h + b_r \\
a^{u}\ &=\ \text{softmax}(W_a h + b_a) \\
t\ &=\ \text{sigmoid} (W_t h + b_t)
\end{aligned}
\end{equation}
where all $W$ and $b$ are weight matrices and bias vectors respectively.

\section{Experiments}
We evaluate BCS-DDQ on a movie-ticket booking task in three settings: simulation, human evaluation and human-in-the-loop training. All the experiments are conducted on the text level.

\subsection{Setup}
\paragraph{Dataset.} 
The dialogue dataset used in this study is a subset of the movie-ticket booking dialogue dataset released in Microsoft Dialogue Challenge~\cite{li2018microsoft}. Our dataset consists of 280 dialogues, which have been manually labeled based on the schema defined by domain experts, as in Table~\ref{tab:schema_annot}. The average length of these dialogues is 11 turns. 

\begin{table}[htbp]
\centering
\small
\begin{tabular}{cllll}
\hline
\hline
\multirow{3}[2]{*}{Intent} & \multicolumn{4}{l}{request, inform, deny, confirm\_question,} \\
& \multicolumn{4}{l}{confirm\_answer, greeting, closing, not\_sure,} \\
& \multicolumn{4}{l}{multiple\_choice, thanks, welcome} \\
\hline
\multirow{4}[2]{*}{Slot} & \multicolumn{4}{l}{city, closing, date, distanceconstraints,} \\
& \multicolumn{4}{l}{greeting, moviename, numberofpeople,} \\
& \multicolumn{4}{l}{price, starttime, state, taskcomplete, theater,} \\
& \multicolumn{4}{l}{theater\_chain, ticket, video\_format, zip} \\
\hline
\hline
\end{tabular}%
\vspace{-2mm}
\caption{The dialogue annotation schema}
\label{tab:schema_annot}%
\end{table}%

\paragraph{Dialogue Agents.} 
We benchmark the BCS-DDQ agent with several baseline agents:
\begin{itemize}[noitemsep,leftmargin=*,topsep=0pt]
\item The \textbf{SL} agent is learned by a variant of imitation learning
\cite{lipton2016efficient}. At the beginning of training, the entire budget is used to collect human-human dialogues, based on which the dialogue agent is trained.
\item The \textbf{DQN} agent is learned by standard DQN 
At each epoch of training, the budget is spent on human-agent interactions, and the agent is trained by direct RL.
\item The \textbf{DDQ} agent is learned by the DDQ method \cite{Peng2018IntegratingPF}. The training process is similar to that of the DQN agent, differing in that DDQ integrates a jointly-trained world model to generate simulated experience which can further improve the dialogue policy. At each epoch of training, the budget is spent on human-agent interactions.
\item The \textbf{BCS-DDQ} agent is learned by the proposed BCS-DDQ approach. For a fair comparison, we use the same number of training epochs $m$ used for the DQN and DDQ agents.
\end{itemize}

\paragraph{Hyper-parameter Settings.}
We use an MLP to parameterize function $Q(\cdot)$ in all the dialogue agents (SL, DQN, DDQ and BCS-DDQ), with hidden layer size set to 80. The $\epsilon$-greedy policy is adopted for exploration. We set discount factor $\gamma = 0.9$. The target value function $Q'(\cdot)$ is updated at the end of each epoch. The world model contains one shared hidden layer and three task-specific hidden layers, all of size 80. The number of planning steps is set to 5 for using the world model to improve the agent's policy in DDQ and BCS-DDQ frameworks. Each dialogue is allowed a maximum of 40 turns, and dialogues exceeding this maximum are considered failures. 
Other parameters used in BCS-DDQ are set as $l=128, d=10$.

\paragraph{Training Details.}
The parameters of all neural networks are initialized using a normal distribution with a mean of 0 and a variance of $\sqrt{6/(d_{row}+d_{col})}$, where $d_{row}$ and $d_{col}$ are the number of rows and columns in the structure~\cite{glorot2010understanding}.
All models are optimized by RMSProp~\cite{tieleman2012lecture}. The mini-batch size is set to 16 and the initial learning rate is 5e-4. The buffer sizes of $B^{r}$ and $B^{s}$ are set to 3000. In order to train the agents more efficiently, we utilized a variant of imitation learning, Reply Buffer Spiking \cite{lipton2016efficient}, to pre-train all agent variants at the starting stage. 

\subsection{Simulation Evaluation}
In this setting, the dialogue agents are trained and evaluated by interacting with the user simulators \cite{li2016user} instead of real users. In spite of the discrepancy between simulated and real users, this setting enables us to perform a detailed analysis of all agents without any real-world cost.
During training, the simulator provides a simulated user response on each turn and a reward signal at the end of the dialogue. The dialogue is considered successful if and only if a movie ticket is booked successfully and the information provided by the agent satisfies all the user's constraints (user goal). When the dialogue is completed, the agent receives a positive reward of $2*L$ for success, or a negative reward of $-L$ for failure, where $L$ is the maximum number of turns allowed (40). To encourage shorter dialogues, the agent receives a reward of $-1$ on each turn.

In addition to the user simulator, the training of SL and BCS-DDQ agents requires a high-performance dialogue agent to play the role of the human agent in collecting human-human conversational data. In the simulation setting, we leverage a well-trained DQN agent as the human agent.

\begin{figure}[t]
\centering
\includegraphics[width=\linewidth]{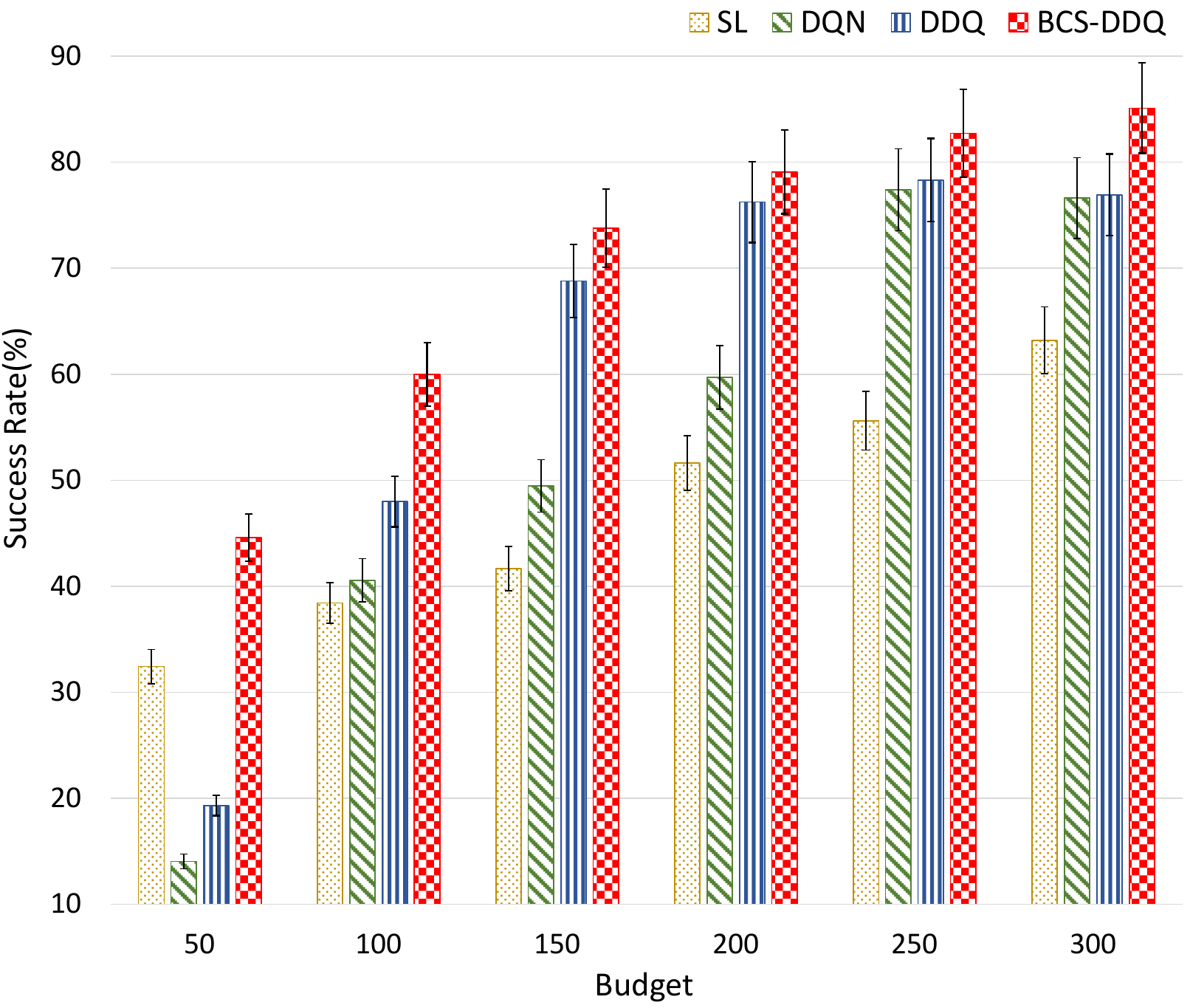}
\caption{The success rates of different agents (SL, DQN, DDQ, BCS-DDQ) given a fixed budget ($b= \{50, 100, 150, 200, 250, 300\}$). Each number is averaged over 5 runs, each run tested on 50 dialogues.}
\label{fig:sim_all_budgets}
\end{figure}

\paragraph{Main Results.}
We evaluate the performance of all agents (SL, DQN, DDQ, BCS-DDQ) given a fixed budget ($b = \{50, 100, 150, 200, 250, 300\}$). As shown in Figure \ref{fig:sim_all_budgets}, the BCS-DDQ agent consistently outperforms other baseline agents by a statistically significant margin. Specifically, when the budget is small ($b = 50$), SL does better than DQN and DDQ
that haven't been trained long enough to obtain significant positive reward. BCS-DDQ leverages human demonstrations to explicitly guide the agent's learning when the agent's performance is very bad. In this way, BCS-DDQ not only takes advantage of imitation learning, but also further improves the performance via exploration and RL. As the budget increases, DDQ can leverage real experiences to learn a good policy. Our method achieves better performance than DDQ, demonstrating that the BCS module can better utilize the budget by directing exploration to parts of the state-action space that have been less explored.

\begin{figure}[t]
\centering
\includegraphics[width=\linewidth]{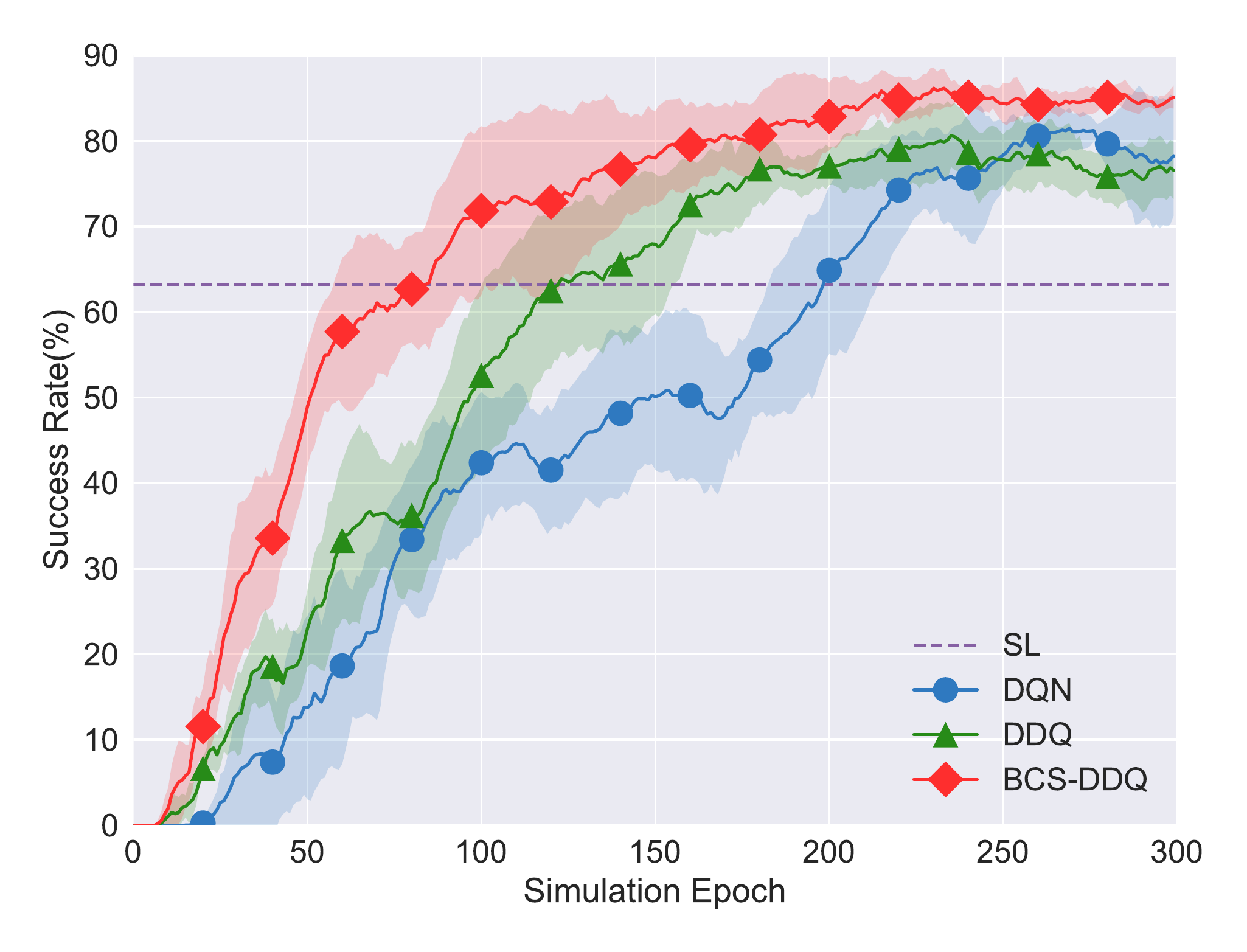}
\caption{The learning curves of different agents (DQN, DDQ and BCS-DDQ) with budget $b=300$.}
\label{fig:sim_curves_budget_300}
\end{figure}

\paragraph{Learning Curves.}
We also investigate the training process of different agents. Figure \ref{fig:sim_curves_budget_300} shows the learning curves of different agents with a fixed budget ($b=300$). At the beginning of training, similar to a very small budget situation, the performance of the BCS-DDQ agent improves faster thanks to its combination of imitation learning and reinforcement learning. After that, BCS-DDQ consistently outperforms DQN and DDQ as training progresses. This proves that the BCS module can generate higher quality dialogue experiences for training dialogue policy.

\begin{table*}[t]
\centering
\small
\begin{tabular}{l|c|c|c|c|c|c|c|c|c}
\hline
\multirow{2}{*}{Agent} & \multicolumn{3}{c|}{Epoch=100} & \multicolumn{3}{c|}{Epoch=150} & \multicolumn{3}{c}{Epoch=200} \\ 
\cline{2-10} 
         & Success   & Reward   & Turns   & Success   & Reward   & Turns   & Success   & Reward   & Turns  \\ 
\hline 
\hline
DQN      & 0.3032    & -18.77   & 32.31   & 0.4675    & 2.07     & 30.07   & 0.5401    & 18.94    & 26.59  \\ \hline
DDQ      & 0.4204    & -2.24    & 27.34   & 0.5467    & 15.46    & 22.26   & 0.6694    & 32.00    & 18.66  \\ \hline
BCS-DDQ  & \textbf{\textcolor{blue}{0.7542}}    & 43.80    & 15.42   & \textbf{\textcolor{blue}{0.7870}} & 47.38    & 16.13   & \textbf{\textcolor{blue}{0.7629}}    & 44.45    & 16.20 \\
\hline
\end{tabular}
\caption{The performance of different agents at training epoch = \{100, 150, 200\} in the human-in-the-loop experiments. The differences between the results of all agent pairs evaluated at the same epoch are statistically significant ($p < 0.05$). (Success: success rate)}
\label{table:result_human_in_loop}
\end{table*}

\subsection{Human Evaluation}
For human evaluation, real users interact with different agents without knowing which agent is behind the system. At the beginning of each dialogue session, we randomly pick one agent to converse with the user. The user is provided with a randomly-sampled user goal, and the dialogue session can be terminated at any time, if the user believes that the dialogue is unlikely to succeed, or if it lasts too long. In either case, the dialogue is considered as failure. At the end of each dialogue, the user is asked to give explicit feedback about whether the conversation is successful.

\begin{figure}[t]
\centering
\includegraphics[width=\linewidth]{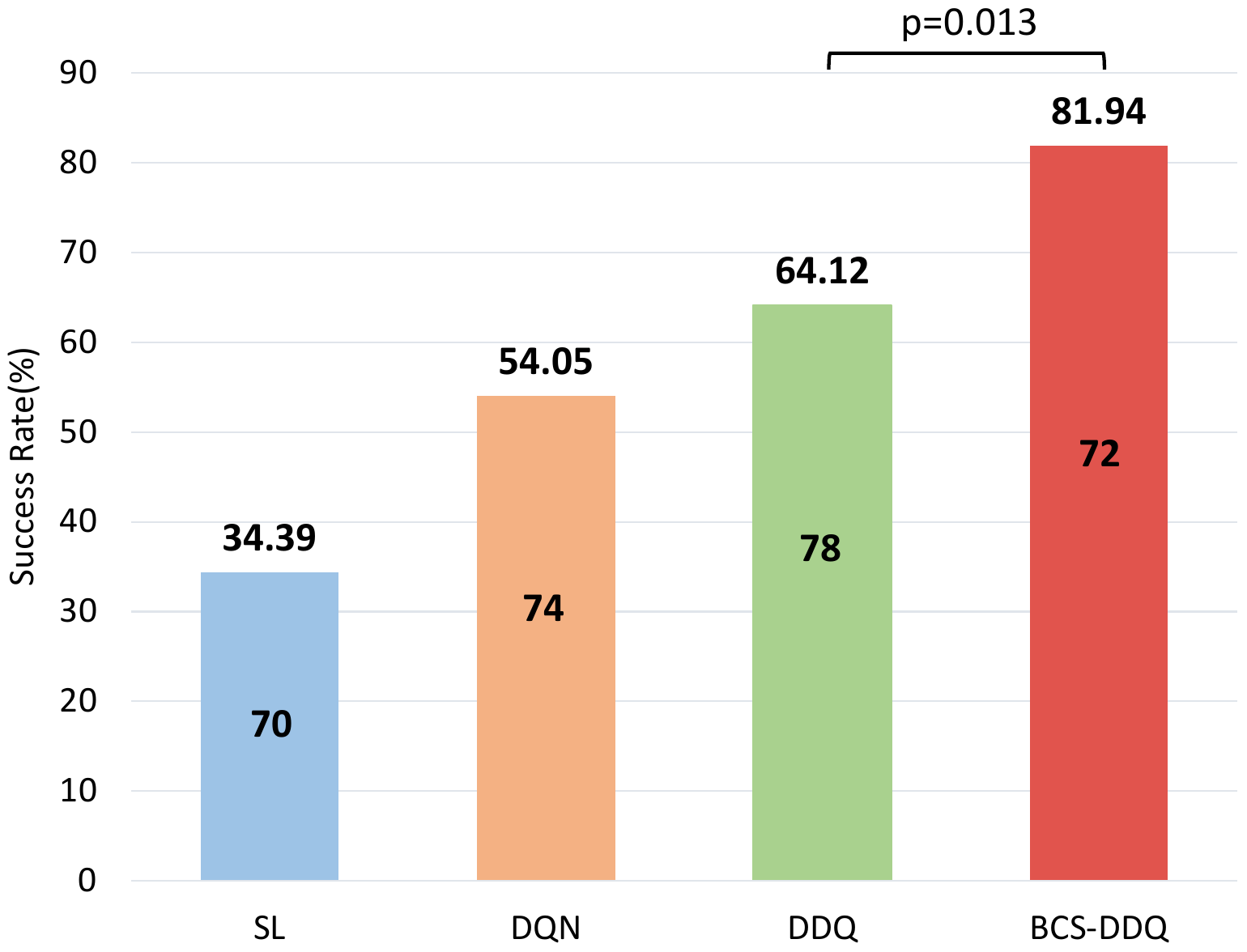}
\caption{The human evaluation results for SL, DQN, DDQ and BCS-DDQ agents, the number of test dialogues indicated on each bar, and the p-values from a two-sided permutation test. The differences between the results of all agent pairs are statistically significant ($p < 0.05$).}
\label{fig:human_eval_budget_300}
\end{figure}

Four agents (SL, DQN, DDQ and BCS-DDQ) trained in simulation (with $b=300$) are selected for human evaluation. As illustrated in Figure \ref{fig:human_eval_budget_300}, the results are consistent with those in the simulation evaluation (the rightmost group with budget=300 in Figure~\ref{fig:sim_all_budgets}). In addition, due to the discrepancy between simulated users and real users, the success rates of all agents drop compared to the simulation evaluation, but the performance degradation of BCS-DDQ is minimal. This indicates that our approach is more robust and effective than the others.

\subsection{Human-in-the-Loop Training}

\begin{figure}[t]
\centering
\includegraphics[width=\linewidth]{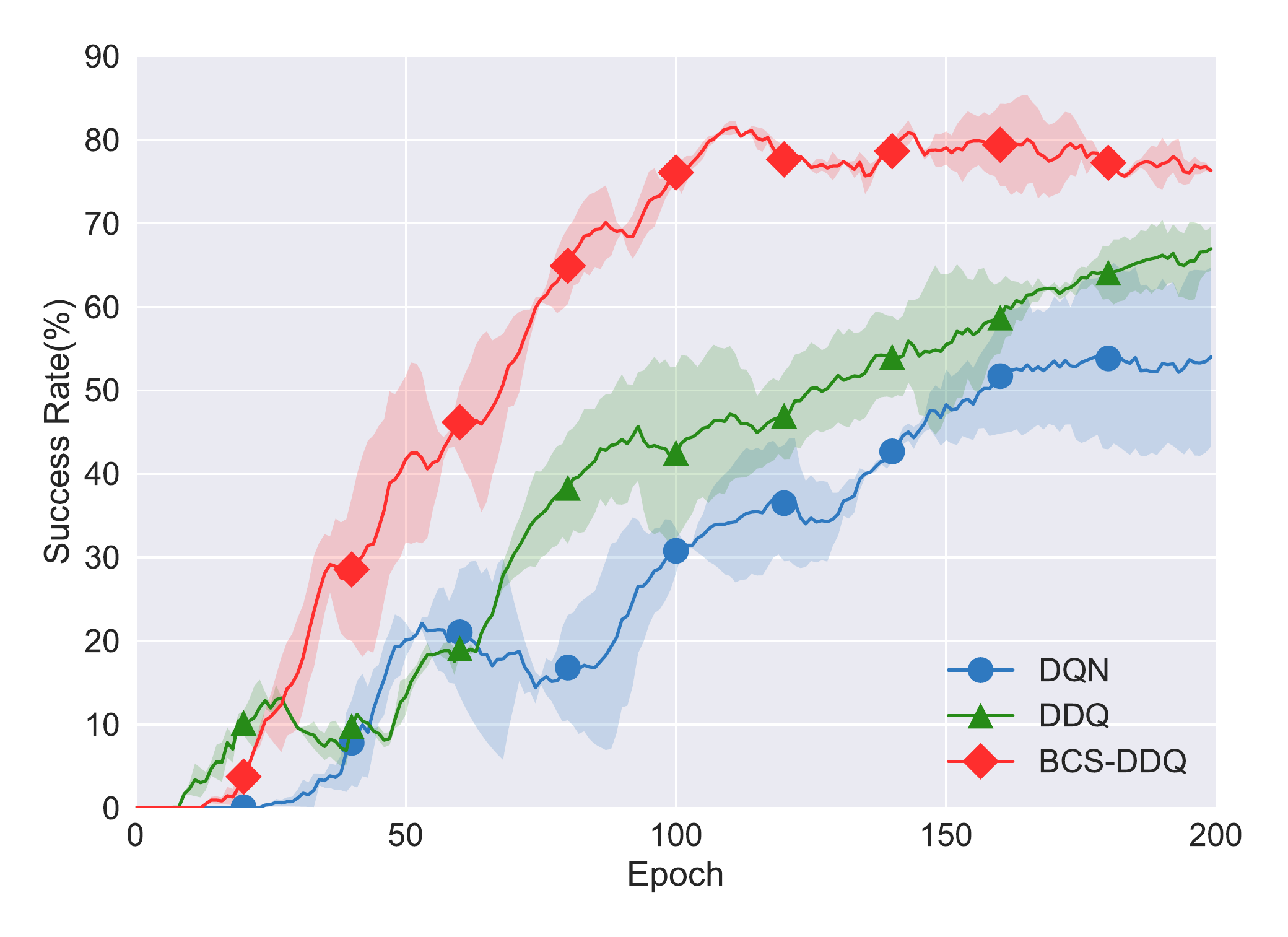}
\caption{Human-in-the-Loop learning curves of different agents with budget $b=200$.}
\label{fig:human_curves_budget_200}
\end{figure}

We further verify the effectiveness of our method in human-in-the-loop training experiments. In this experiment, we replace the user simulator with real users during training. Similar to the human evaluation, based on a randomly-sampled user goal, the user converses with a randomly-selected agent and gives feedback as to whether the conversation is successful. In order to collect human-human conversations during the training of the BCS-DDQ agent, human agents are interacting directly with real users through the dialogue system. 
In a dialogue session, the human agent has access to the conversation history, as well as the current search results from the knowledge base, before selecting each dialogue action in response to the real user.
Each learning curve is trained with two runs, with each run assigning a budget of 200 human interactions.

\begin{table}[!t]
\small
\begin{tabular}{l}
\hline
\multicolumn{1}{c}{\textbf{Real User Dialogue Sample}}  \\
\hline
\begin{tabular}[c]{@{}l@{}}
\textit{movie-ticket} booking user goal:\\
\{\\
\-\hspace{1mm} ``constraint\_slots": \{	 \-\hspace{15mm}     ``request\_slots": \{\\
\-\hspace{3mm}    \textbf{``numberofpeople": ``four"}  \-\hspace{10mm}    ``ticket":``?" \\
\-\hspace{3mm}    \textbf{``moviename": ``creed"}      \-\hspace{15mm}    ``theater": ``?"	\\
\-\hspace{3mm}    \textbf{``city": ``regency"}	 \-\hspace{16mm}   \}\\
\-\hspace{3mm}    \textbf{``date": ``tomorrow"}	         \-\hspace{18mm}    \\
\-\hspace{3mm}    \textbf{``starttime": ``around noon"}	         \-\hspace{18mm}    \\
\-\hspace{1mm}  \} \\
\} \\

\end{tabular} \\
\hline
\multicolumn{1}{c}{\textbf{DDQ agent}} \\

\begin{tabular}[c]{@{}p{4mm}@{}|l@{}} 1 & \textit{usr}: I want to watch creed at around noon.  \end{tabular}\\
\begin{tabular}[c]{@{}p{4mm}@{}|l@{}} 2 & \textit{agt}: around noon is available. \end{tabular} \\
\begin{tabular}[c]{@{}p{4mm}@{}|l@{}} 3 & \textit{usr}: I want four tickets please! \end{tabular} \\
\begin{tabular}[c]{@{}p{4mm}@{}|l@{}} 4 & \textit{agt}: Which city would you like? \end{tabular} \\
\begin{tabular}[c]{@{}p{4mm}@{}|l@{}} 5 & \textit{usr}: I want to watch at regency. \end{tabular} \\
\begin{tabular}[c]{@{}p{4mm}@{}|l@{}} 6 & \textit{agt}: \textcolor{blue}{Which theater would you like? } \end{tabular} \\
\begin{tabular}[c]{@{}p{4mm}@{}|l@{}} 7 & \textit{usr}: \underline{Which theater is available?} \end{tabular} \\
\begin{tabular}[c]{@{}p{4mm}@{}|l@{}} 8 &\textit{agt}: What date would you like to watch it? \end{tabular} \\
\begin{tabular}[c]{@{}p{4mm}@{}|l@{}} 9 &\textit{usr}: I want to set it up tomorrow. \end{tabular} \\   
\begin{tabular}[c]{@{}p{4mm}@{}|l@{}} 10 &\textit{agt}: \textcolor{blue}{Which theater would you like?}  \end{tabular} \\
\begin{tabular}[c]{@{}p{4mm}@{}|l@{}} 11 &\textit{usr}: \underline{Which theater is available?}	\end{tabular} \\
\begin{tabular}[c]{@{}p{4mm}@{}|l@{}} 12 &\textit{agt}: Which city would you like?  \end{tabular} \\
\begin{tabular}[c]{@{}p{4mm}@{}|l@{}} 13 &\textit{usr}: I want to watch at regency. \end{tabular} \\ 
\begin{tabular}[c]{@{}p{4mm}@{}|l@{}} 14 &\textit{agt}: \textcolor{blue}{Which theater would you like?} \end{tabular}\\
\begin{tabular}[c]{@{}p{4mm}@{}|l@{}} 15 &\textit{usr}: \underline{Which theater is available?} \end{tabular} \\
\begin{tabular}[c]{@{}p{4mm}@{}|l@{}} 16 &\textit{agt}: What date would you like to watch it? \end{tabular}  \\
\multicolumn{1}{c}{\textbf{\vdots}}  \\
\hline
\textbf{Failure:} \begin{tabular}[c]{@{}l@{}}Agent does not book the right ticket within the \\ max turns.\end{tabular}
\\ \hline

\multicolumn{1}{c}{\textbf{BCS-DDQ agent}} \\

\begin{tabular}[c]{@{}p{4mm}@{}|l@{}} 1 & \textit{usr}: I want to watch creed at around noon. \end{tabular} \\
\begin{tabular}[c]{@{}p{4mm}@{}|l@{}} 2 & \textit{agt}: Which city would you like?  \end{tabular} \\
\begin{tabular}[c]{@{}p{4mm}@{}|l@{}} 3 & \textit{usr}: I want to watch at regency.  \end{tabular}  \\
\begin{tabular}[c]{@{}p{4mm}@{}|l@{}} 4 & \textit{agt}: What date would you like to watch it?  \end{tabular}  \\
\begin{tabular}[c]{@{}p{4mm}@{}|l@{}} 5 & \textit{usr}: I want to set it up tomorrow.  \end{tabular} \\
\begin{tabular}[c]{@{}p{4mm}@{}|l@{}} 6 & \textit{agt}: \textcolor{blue}{Which theater would you like?} \end{tabular}   \\
\begin{tabular}[c]{@{}p{4mm}@{}|l@{}} 7 & \textit{usr}: \underline{Which theater is available?}  \end{tabular} \\
\begin{tabular}[c]{@{}p{4mm}@{}|l@{}} 8 & \textit{agt}: \textbf{century eastport 16 is available.}  \end{tabular}   \\
\begin{tabular}[c]{@{}p{4mm}@{}|l@{}} 9 & \textit{usr}: Could you help me to book the tickets?  \end{tabular} \\
\begin{tabular}[c]{@{}p{4mm}@{}|l@{}} 8 & \textit{agt}: How many tickets do you need?  \end{tabular}   \\
\begin{tabular}[c]{@{}p{4mm}@{}|l@{}} 9 & \textit{usr}: I want four tickets please!   \end{tabular} \\
\begin{tabular}[c]{@{}p{4mm}@{}|l@{}} 10 & \textit{agt}: \begin{tabular}[c]{@{}l@{}} Great - I was able to purchase 4 tickets for you \\ to see creed tomorrow at century eastport 16 \\ theater in regency at around noon. \end{tabular} \end{tabular} \\
\begin{tabular}[c]{@{}p{4mm}@{}|l@{}} 11 & \textit{usr}: Thank you. \end{tabular}   \\   
\begin{tabular}[c]{@{}p{4mm}@{}|l@{}} 12 &\textit{agt}: Thank you. \end{tabular}   \\   
\hline
\multicolumn{1}{c}{\textbf{Success}}
\\ \hline
\end{tabular}
\small
\centering
\caption{Sample dialogue sessions by DDQ and BCS-DDQ agents trained at epoch 200 (with total budget $b=200$) in the human-in-the-loop experiments: (\textit{agt}: agent, \textit{usr}: user) }
\label{tab:sample_dialogues_comparison}
\end{table}

The main results are presented in Table \ref{table:result_human_in_loop} and Figure \ref{fig:human_curves_budget_200}.
We can see that the BCS-DDQ agent consistently outperforms DQN and DDQ during the course of training, confirming the conclusion drawn from the simulation evaluation. Besides, Table \ref{tab:sample_dialogues_comparison} shows example dialogues produced by two dialogue agents (DDQ and BCS-DDQ) interacting with human users respectively. We can see that DDQ agent fails to respond to the user question ``which theater is available?'', which lead to the repeated inquiry of theater information. By introducing human demonstrations for agent training, BCS-DDQ agent can successfully respond to the available theater information.

\subsection{Ablation Study}
\begin{figure}[t]
\centering
\includegraphics[width=\linewidth]{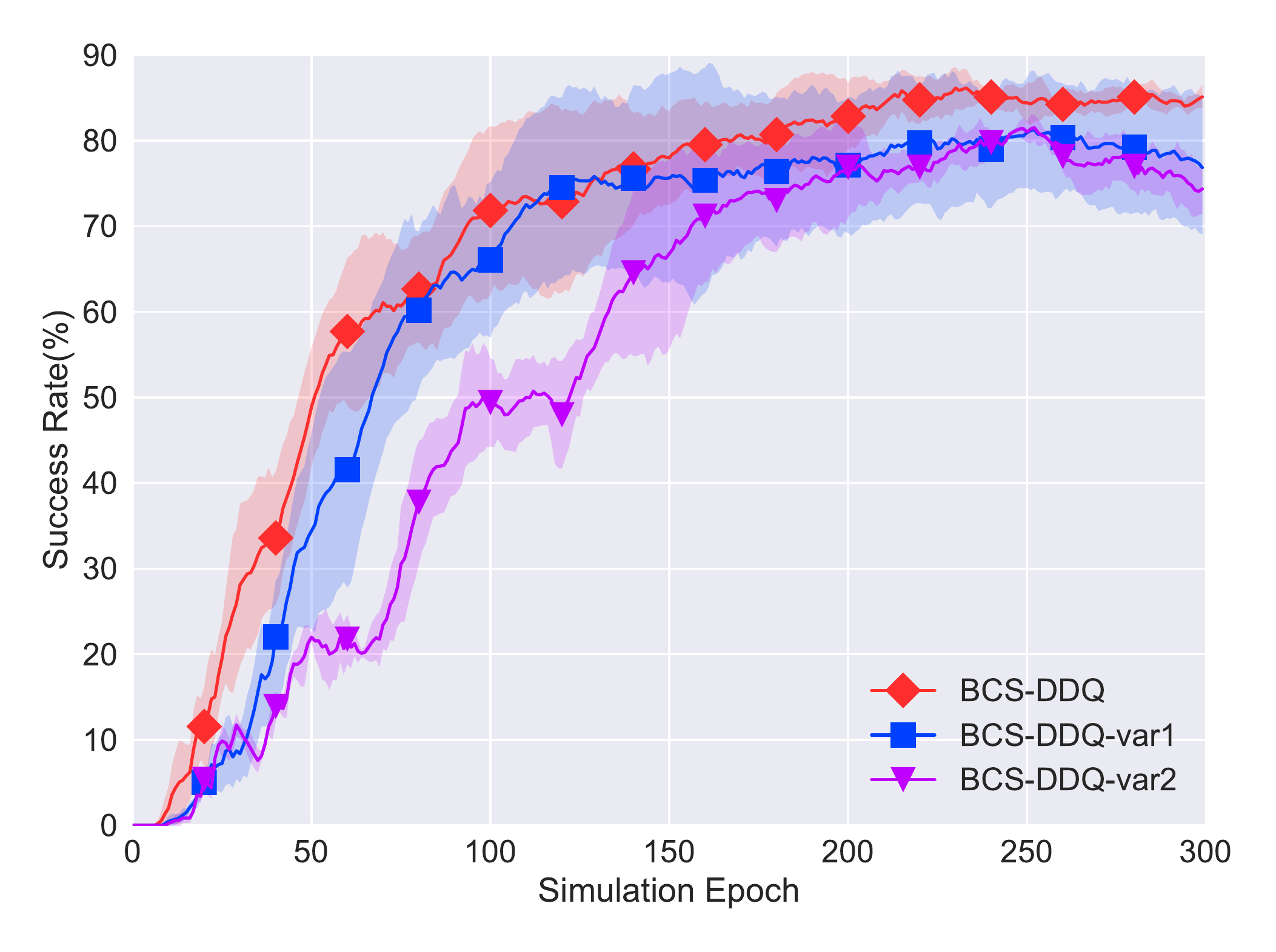}
\caption{The learning curves of BCS-DDQ and its variants agents with budget $b=300$.}
\label{fig:ablation_sim_curves_budget_300}
\end{figure}

We investigate the relative contribution of the budget allocation algorithm and the active sampling strategy in BCS-DDQ by implementing two variant BCS-DDQ agents:
\begin{itemize}[noitemsep,leftmargin=*,topsep=0pt]
\item The \textbf{BCS-DDQ-var1} agent: Replacing the decayed Poisson process with a regular Poisson process in the budget allocation algorithm, which means that the parameter $\lambda$ is set to $\frac{b}{m}$ during training.
\item The \textbf{BCS-DDQ-var2} agent: Further replacing the active sampling strategy with random sampling, based on the BCS-DDQ-var1 agent.
\end{itemize}
The results in Figure \ref{fig:ablation_sim_curves_budget_300} shows that the budget allocation algorithm and active sampling strategy are helpful for improving a dialogue policy in the limited budget setting. The active sampling strategy is more important, without which the performance drops significantly. 

\section{Conclusion}
We presented a new framework BCS-DDQ for task-oriented dialogue policy learning. Compared to previous work, our approach can better utilize the limited real user interactions in a more efficient way in the fixed budget setting, and its effectiveness was demonstrated in the simulation evaluation, human evaluation, including human-in-the-loop experiments. 

In future, we plan to investigate the effectiveness of our method on more complex task-oriented dialogue datasets. Another interesting direction is to design a trainable budget scheduler. In this paper, the budget scheduler was created independently of the dialogue policy training algorithm, so a trainable budget scheduler may incur additional cost. One possible solution is transfer learning, in which we train the budget scheduler on some well-defined dialogue tasks, then leverage this scheduler to guide the policy learning on other complex dialogue tasks.


\section{Acknowledgments}
We appreciate 
Sungjin Lee,
Jinchao Li,
Jingjing Liu, 
Xiaodong Liu, 
and Ricky Loynd 
for the fruitful discussions. We would like to thank the volunteers from Microsoft Research for helping us with the human evaluation and the human-in-the-loop experiments. We also thank the anonymous reviewers for their careful reading of our paper and insightful comments. This work was done when Zhirui Zhang was an intern at Microsoft Research.

\bibliography{acl2019}
\bibliographystyle{acl_natbib}

\appendix

\end{document}